\documentclass{article}
\usepackage{spconf,amsmath,graphicx}
\usepackage{times}
\usepackage{epsfig}
\usepackage{amsmath}
\usepackage{amssymb}
\usepackage{mathtools}
\usepackage{multirow}
\usepackage{xcolor}
\usepackage{xurl}
\usepackage{makecell,rotating}
\usepackage[labelsep=period]{caption}
\newcommand\blfootnote[1]{%
	\begingroup
	\renewcommand\thefootnote{}\footnote{#1}%
	\addtocounter{footnote}{-1}%
	\endgroup
}

\usepackage{bbding}
\usepackage{amsmath}
\usepackage{amsfonts}
\usepackage{algorithm}
\usepackage{algorithmic}
\usepackage{caption}
\usepackage{subfigure}
\usepackage{graphicx}

\title{Visual Answer Localization with Cross-modal \\ Mutual Knowledge Transfer}
\name{Yixuan Weng*$^\dag$, Bin Li*$^{\star\ddagger}$}
\address{ $^\dag$ National Laboratory of Pattern Recognition Institute of Automation, Chinese Academy Sciences \\
	$^\ddagger$	College of Electrical and Information Engineering, Hunan University }
\begin{document}
	%
	
	\maketitle
	
	%
	%
	\begin{abstract}
		%
		%
		%
		%
		%
	The goal of visual answering localization (VAL) in the video is to obtain a relevant and concise time clip from a video as the answer to the given natural language question. Early methods are based on the interaction modelling between video and text to predict the visual answer by the visual predictor. Later, using the textual predictor with subtitles for the VAL proves to be more precise. However, these existing methods still have cross-modal knowledge deviations from visual frames or textual subtitles. In this paper, we propose a cross-modal mutual knowledge transfer span localization (MutualSL) method to reduce the knowledge deviation. MutualSL has both visual predictor and textual predictor, where we expect the prediction results of these both to be consistent, so as to promote semantic knowledge understanding between cross-modalities. On this basis, we design a one-way dynamic loss function to dynamically adjust the proportion of knowledge transfer. We have conducted extensive experiments on three public datasets for evaluation. The experimental results show that our method outperforms other competitive state-of-the-art (SOTA) methods, demonstrating its effectiveness\footnote{All the experimental datasets and codes are open-sourced on the website {https://github.com/WENGSYX/MutualSL}.}.
	\end{abstract}
	\begin{keywords}
		Cross-modal, Mutual Knowledge Transfer, Visual Answer Localization
		\blfootnote{*: These authors contributed equally to this work.}
		\blfootnote{$^{\star}$: Corresponding author.}
	\end{keywords}
		\vspace{-0.2cm}
	\section{Introduction}
	\begin{figure}[t]
		\centering
		\includegraphics[width=0.95\linewidth]{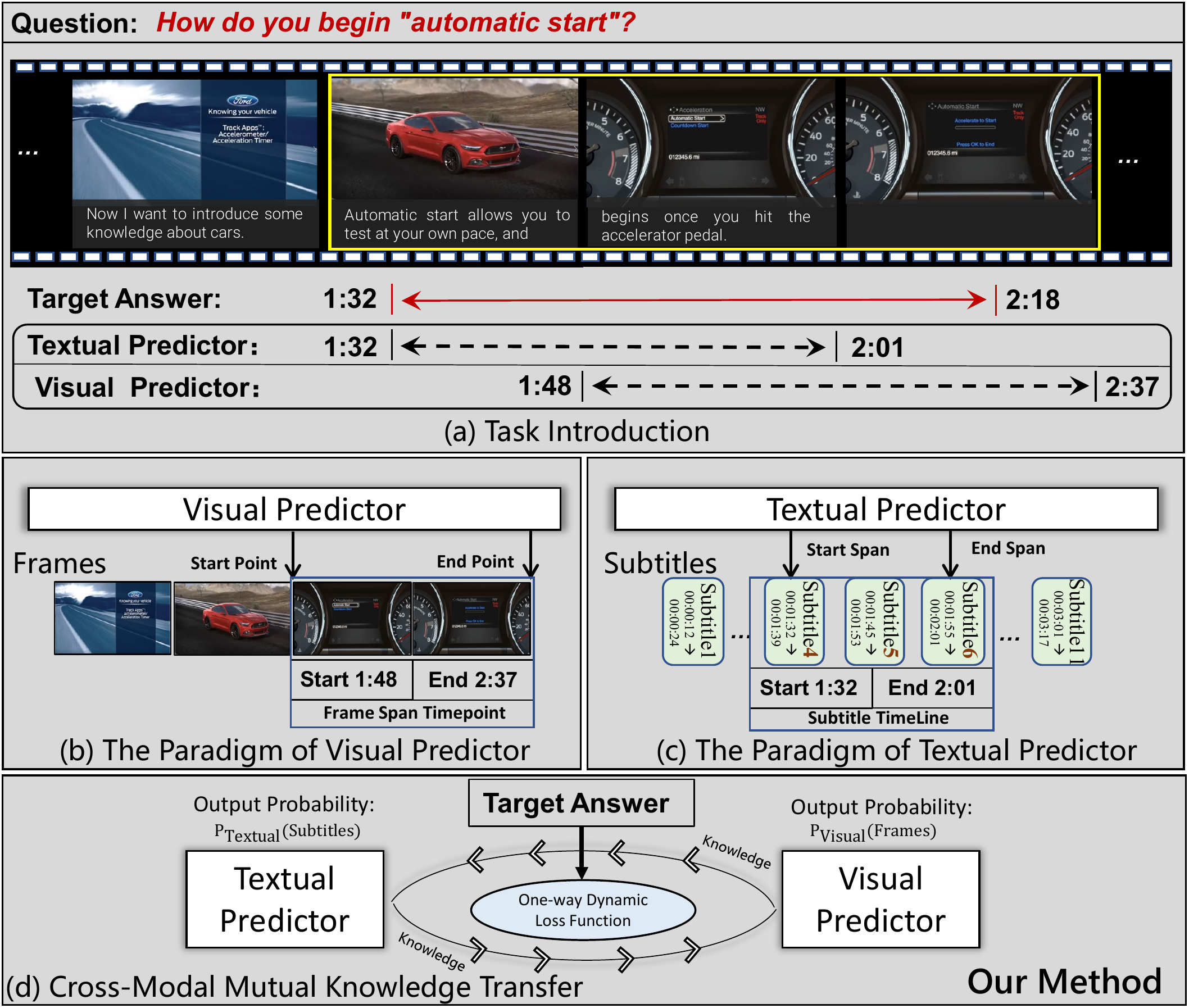}
		\vspace{-0.2cm}
		\caption{Task description of the visual answer localization, where the below is the paradigms of the previous methods and our method.
		}
		\vspace{-0.6cm}
		\label{sample2}
	\end{figure}
	\label{sec: Intro}
	\begin{figure*}[t]
		\centering
		\includegraphics[width=15.6cm]{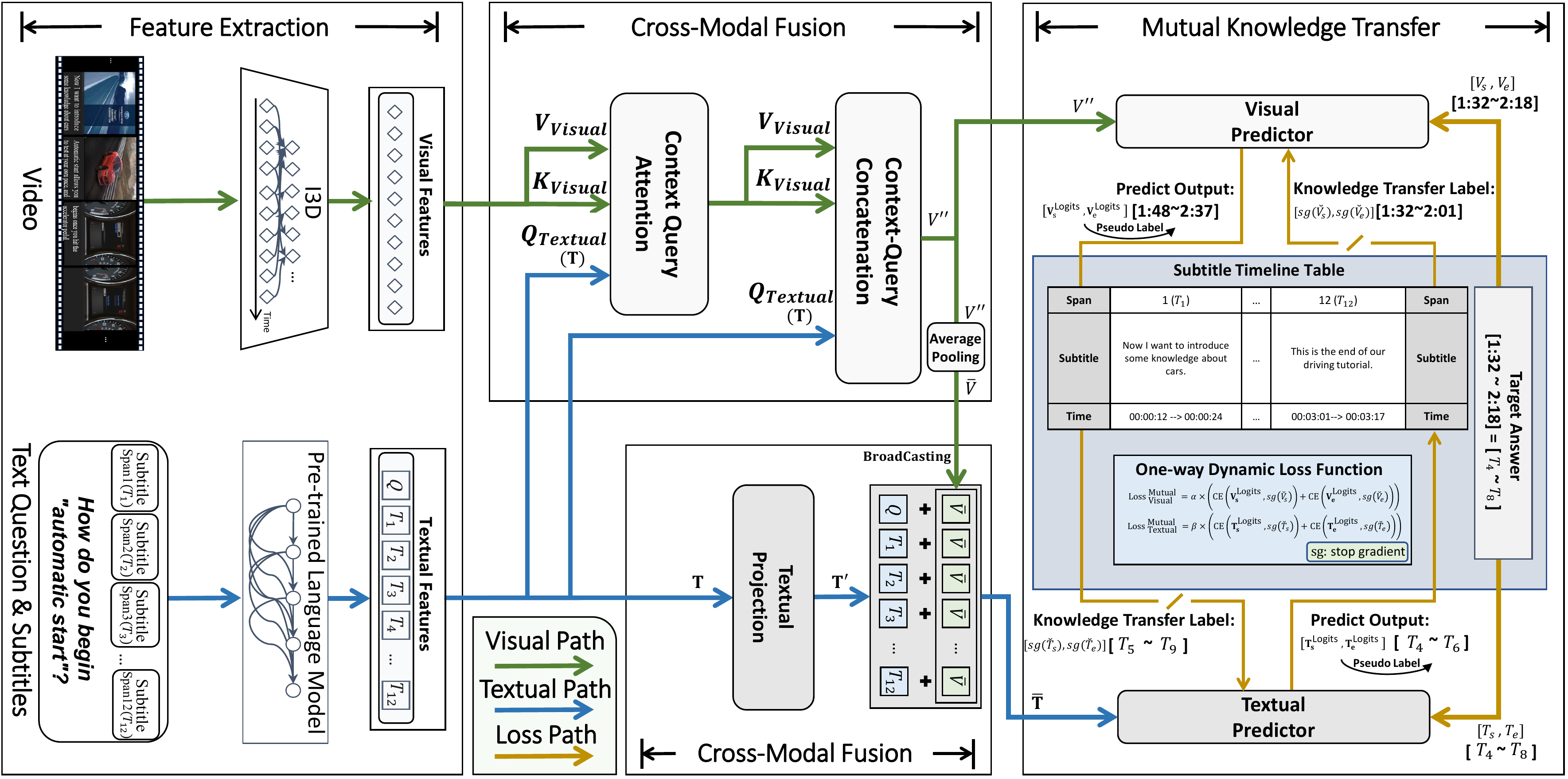}
		\vspace{-0.2cm}
		\caption{Overview of the proposed cross-modal mutual knowledge transfer span localization (MutualSL).}
		\label{framework}
		\vspace{-0.7cm}
	\end{figure*}
	The explosion of online videos has changed the way that people obtain information, and knowledge \cite{SalmanKhan2022TransformersIV, xu2022transformers}. Various video platforms make it more convenient for people to perform video queries \cite{koupaee2018wikihow, torhonen2019fame}. However, people who want to get direct instructions or tutorials from the video often need to browse the video content several times to locate relevant parts, which usually takes time and effort \cite{AnthonyColas2019TutorialVQAQA}.
	Visual answer localization (VAL) is an emerging technology to solve the above problem \cite{gupta2022dataset}, and has received wide attention because of its practical value \cite{gupta-demner-fushman-2022-overview,kusa-etal-2022-dossier}. As shown in Fig.~\ref{sample2}(a), the task of VAL is to find a time clip that can answer the given question. For example, when inputting ``How do you begin `automatic start'{''}, you may need to find a clip according to voice content (or transcribed text subtitles) and visual frames. The VAL technology can not only recognize the relevant video clips to the text questions but also return the target visual answer (1:32 \textasciitilde~2:18).
	\par
	The existing VAL method can be mainly divided into visual predictor and textual predictor according to the prediction contents. The paradigm of visual predictor is shown in Fig.~\ref{sample2}(b). The video information is first extracted according to the frame features, and then these frame features queried by the question are used to predict the relevant time points \cite{zhang2020span, tang2021frame}. The paradigm of textual predictor is shown in Fig.~\ref{sample2}(c). The textual predictor adopts a span-based method to model the cross-modal information, where the predicted span intervals with subtitle timeline are used as the final results \cite{kusa-etal-2022-dossier, li2022towards}. 
	\par
	The performance of the textual predictor is better than the visual one \cite{gupta-demner-fushman-2022-overview}, because it uses the additional subtitle information, and embeds visual information into the text feature space with the visual information as an auxiliary feature. However, as shown in Fig.~\ref{sample2}(a), results from both two predictors suffer cross-modal knowledge deviations. For the textual predictor, if the video lacks subtitle information for a long clip, this clip cannot be located; For the visual predictor, it is difficult to have continuous clip prediction because of the frequent changing of video scenes and semantics to the question.
	\par
	In this paper, we propose a novel cross-modal mutual knowledge transfer span localization (MutualSL) method to reduce the cross-modal knowledge deviation shown in Fig.~\ref{sample2}(d). Specifically, the MutualSL uses both visual predictor and textual predictor, where these two predictors have different prediction targets so that they have different strength perceptions of different-modal information. We expect that these two predictors can enhance the information perception of their own modal. Each predictor needs to predict the output value of another predictor on the basis of the target answer in the training stage. Then we design a one-way dynamic loss function (ODL) to dynamically adjust the knowledge transfer, which can alleviate the difference of cross-modal knowledge transferring in the training process.
	\par
	Our contributions are as follows: (1) we propose the MutualSL method, which for the first time uses two different predictors in VAL tasks meanwhile, and uses a Look-up Table to achieve cross-modal knowledge transfer; (2) We design ODL to dynamically adjust the knowledge transfer, which can alleviate the differences in knowledge transfer between different predictors; (3) We have conducted extensive experiments to prove the effectiveness of the MutualSL, where results show that the proposed method outperforms all other competitive SOTA methods in VAL tasks.
		\vspace{-0.3cm}
	\section{Method}\label{section1: selfmodel}
	\vspace{-0.1cm}
	\subsection{Task Definition}
	
	Given an untrimmed video $V$ with a duration of $k$ seconds, the corresponding subtitle  $S=\{T_i\}_{i=1}^r$ and the text question is $Q$, the VAL task requires us to predict the most relevant visual clips within the video $[V_s^{*}, V_e^{*}]  \subseteq   V$ that answer the question $Q$, where $T_i$ is the subtitle of each span, $r$ is the subtitle span length, and $[V_s, V_e]$ is defined as the target time clip answer, ${s,e \in [1,k]}$. Moreover, it provides a subtitle timeline table, which is translated a span into corresponding timeline span from each subtitle set $S$. We can use the subtitle timeline table as the Look-up Table, such as providing accurate target answers for textual predictor transferred from the frame span timepoints, and {vice versa}.
		\vspace{-0.1cm}
	\begin{equation}
		[V_s^{*},V_e^{*}] = \mathop{\mathrm{Argmin}}\limits_{V_s, V_e}(P([V_s,V_e]|V,S,Q))
	\end{equation}
	\vspace{-0.8cm}
	\subsection{Main Structure}
	As shown in Fig.~\ref{framework}, the MutualSL is divided into three parts, which are Feature Extraction, Cross-modal Fusion, and Mutual Knowledge Transfer. The Mutual Knowledge Transfer includes visual predictor and textual predictor.
	\par\noindent\textbf{Feature Extraction.} Following the previous method \cite{zhang2020span,li2022towards}, we use the pre-trained visual model I3D \cite{carreira2017quo} and the pre-trained language model (PLM) \cite{he2021debertav3} to extract feature vectors from video ${V}$ and concatenated texts $T = [Q, T_1,\ldots, T_r]$ respectively. These pre-trained models can provide us with high-quality information representation.
			\vspace{-0.1cm}
	\begin{equation}
		\mathbf{V} = \mathrm{I3D}(V) , \mathbf{T} = \mathrm{PLM}(T) ,
	\end{equation}
	\noindent where $\mathbf{V} \in \mathbb{R}^{k\times d}$ and $\mathbf{T} \in \mathbb{R}^{n\times d}$, the $d$ is the dimension and $n$ is the length of the concatenated text tokens of T.
	\par \noindent\textbf{Cross-modal Fusion.} We use context query attention (CQA) \cite{zhang2020span} to capture the cross-modal interaction between visual and textual to enhance the semantics in the visual path. CQA adopts two attention mechanism context to query ($\mathcal{D}$) and query to context ($\mathcal{F}$) processes for cross-modal modeling, where $\mathcal{G}_{r} \in \mathbb{R}^{k \times n}$ and $\mathcal{G}_{c} \in \mathbb{R}^{k \times n}$ represent the row- and column-wise normalization of $\mathcal{G}$ by SoftMax.
	\begin{equation}
		\mathcal{D}=\mathcal{G}_{r}\cdot\mathbf{{T}}\in\mathbb{R}^{k\times d}, \mathcal{F}=\mathcal{G}_{c}\cdot\mathcal{G}_{r}^{T}\cdot\mathbf{{V}}\in\mathbb{R}^{k\times d}\nonumber
	\end{equation}
	We use one layer of feedforward neural network ($\mathrm{FFN_C}$) and convolution layer (in\_channels = $2d$, out\_channels = $d$) as Context-Query Concatenation to capture deeper semantic information, where \{$\mathbf{{V^{\prime}}}$, $\mathbf{{V^{\prime\prime}}}$\}$\in\mathbb{R}^{k\times d}$.
	\begin{equation}
		\mathbf{ V^{\prime}}=\mathrm{FFN_C}\big([\mathbf{{V}};\mathcal{D};\mathbf{{V}}\odot\mathcal{D};\mathbf{{V}}\odot\mathcal{F}]\big)
	\end{equation}
	\begin{equation}
		\mathbf{ V^{\prime\prime}}={\text{Conv1d}}\big(\text{Concat}[\text{Attention}(\mathbf{{V^{\prime}}},\mathbf{T});\mathbf{T}]\big)
	\end{equation}
	
	We use a textual projection layer ($\mathrm{FFN_P}$) to extract text features $\mathbf{ T^{\prime}} \in\mathbb{R}^{n\times d}$ in the text path. Then we embed the average pooled visual feature $\overline{\mathbf{V}}$ into each token $T_j$ in $\mathbf{ T^{\prime}}$ through the broadcast mechanism, where $\overline{\mathbf{V}} \in \mathbb{R}^{d}$ and $T_j \in \mathbb{R}^{1\times d}$,
	\begin{equation}
		\mathbf{ T^{\prime}}= \mathrm{FFN_P}(\mathbf{T}),\overline{\mathbf{V}} = \mathrm{AvgPool(\mathbf{V^{\prime\prime}})})
	\end{equation}
	\begin{equation}
		\overline{\mathbf{T}}= \{\overline{\mathbf{V}}+T_j\}_{j=1}^n, \overline{\mathbf{T}}\in \mathbb{R}^{n\times d}
	\end{equation}
	\textbf{Visual Predictor.} We use two unidirectional LSTMs, including  $\mathrm{LSTM_{Start}}$ and $\mathrm{LSTM_{End}}$, where the in\_channels = $k\times d$, out\_channels = $k\times d$.
	Two feedforward layers $\mathrm{FFN_{Start}^{Visual}}$ and $\mathrm{FFN_{End}^{Visual}}$( in\_channels = $k\times d$, out\_channels = $k$) are adopted to construct a visual span predictor. We input the features $\mathbf{V^{\prime\prime}}$ into LSTMs, then use the feedforward layer to calculate the predicted time point logits, including the start time point and end time point.
	\begin{equation}
		\mathbf{V_s^{Logits}} = \mathrm{FFN_{Start}^{Visual}}(\mathrm{LSTM_{Start}}(\mathbf{V^{\prime \prime}}))
	\end{equation}
	\begin{equation}
		\mathbf{V_e^{Logits}} = \mathrm{FFN_{End}^{Visual}}(\mathrm{LSTM_{End}}(\mathbf{V^{\prime\prime}}))
	\end{equation}
	\textbf{Textual Predictor.} We follow to the structure of QANet \cite{AdamsWeiYu2018QANetCL} and calculate the probability of outputting the start and the end subtitle point through two different feedforward layers $\mathrm{FFN_{Start}^{Textual}}$ and $\mathrm{FFN_{End}^{Textual}}$, where the in\_channels = $n\times d$, out\_channels = $n$.
	\begin{equation}
		\mathbf{T_s^{Logits}} =\mathrm{FFN_{Start}^{Textual}}(\overline{\mathbf{T}}),
		\mathbf{T_e^{Logits}} =\mathrm{FFN_{End}^{Textual}}(\overline{\mathbf{T}})
	\end{equation}
	\textbf{Loss Function.} We adopt Cross-Entropy (CE) function to maximize the visual predictor logits of the target span-point $[V_s,V_e]$. Also, we convert $[V_s, V_e]$ to $[T_s, T_e]$ by subtitle timeline Look-up Table for the loss calculation.
	\begin{equation}
		\mathrm{Loss_{Visual}} = \mathrm{CE}(\mathbf{V_s^{Logits}},V_s)+\mathrm{CE}(\mathbf{V_e^{Logits}},V_e)
	\end{equation}
	\begin{equation}
		\mathrm{Loss_{Textual}} = \mathrm{CE}(\mathbf{T_s^{Logits}},T_s)+\mathrm{CE}(\mathbf{T_e^{Logits}},T_e)
	\end{equation}
	
	\begin{table*}[t]
		\centering
		\caption{Performance on three public datasets compared with several SOTA methods.}
		\vspace{-0.2cm}
		\resizebox{1\textwidth}{!}{
			\begin{tabular}{c|cccc|cccc|cccc}
				\noalign{\hrule height 1pt}
				\multicolumn{1}{c|}{\thead{\multirow{2}{*}{\begin{tabular}[c]{@{}c@{}} \normalsize {Method} \end{tabular}}}}&\multicolumn{4}{c|}{MedVidQA}&\multicolumn{4}{c|}{TutorialVQA}&\multicolumn{4}{c}{VehicleVQA}    \\ 
				&IoU=0.3&IoU=0.5&IoU=0.7&mIoU& IoU=0.3&IoU=0.5&IoU=0.7&mIoU& IoU=0.3&IoU=0.5&IoU=0.7&mIoU \\ \hline
				VSLBase \cite{zhang2020span}&27.66& 14.19&6.99&21.01&10.84&9.58&0.37&8.71&18.95&8.64&4.28&20.11\\
				TMLGA \cite{CristianRodriguezOpazo2020ProposalfreeTM}&23.87&14.84&6.21&20.49&-&-&-&-&-&-&-&-\\
				VSLNet \cite{zhang2020span}&30.32&16.61&8.39&22.41&9.96&9.21&0.00&8.58&16.53&8.47&4.03&20.07 \\
				ACRM \cite{tang2021frame}&24.83&16.55&10.96&22.89&12.61&5.17&1.26&11.18&20.77&12.10&8.27&22.28\\
				RaNet \cite{JialinGao2021RelationawareVR}&32.90&20.64&15.48&27.48&-&-&-&-&-&-&-&-\\
				MoR \cite{kusa-etal-2022-dossier}&47.10&22.74&10.97&30.67&-&-&-&-&-&-&-&-\\
				VPTSL \cite{li2022towards}&77.42&\textbf{61.94}&\textbf{44.52}&57.81&50.07&40.01&25.79&40.20&74.15&67.15&54.59&64.51\\ \hline

				\multicolumn{1}{c|}{MutualSL}&\textbf{80.65}&\textbf {61.94}&39.99&\textbf {58.32}&\textbf {60.14}&\textbf {43.59}&\textbf {28.28}&\textbf {43.48}&\textbf {78.74}&\textbf {69.81}&\textbf {53.14}&\textbf {65.74} \\ 
				\noalign{\hrule height 1pt}

		\end{tabular}}
		\vspace{-0.2cm}
		
		\label{t11}
	\end{table*}

	\begin{table*}[t]
		\centering
		\caption{ {{We report the effect of whether to conduct cross-modal mutual knowledge transfer for Visual Predictor (VP) and Textual Predictor (TP) respectively. Both predictors are output from MutualSL, but their VAL performance is different. Therefore, we report the results on whether using Mutual Knowledge Transfer (MKT).}}}
		\vspace{-0.2cm}
		\resizebox{1\textwidth}{!}{
			\begin{tabular}{c|cccc|cccc|cccc}
				\noalign{\hrule height 1pt}
				\multicolumn{1}{c|}{\thead{\multirow{2}{*}{\begin{tabular}[c]{@{}c@{}} \normalsize {Method} \end{tabular}}}}&\multicolumn{4}{c|}{MedVidQA}&\multicolumn{4}{c|}{TutorialVQA}&\multicolumn{4}{c}{VehicleVQA}    \\ 
				&IoU=0.3&IoU=0.5&IoU=0.7&mIoU& IoU=0.3&IoU=0.5&IoU=0.7&mIoU& IoU=0.3&IoU=0.5&IoU=0.7&mIoU \\  \hline
				
				Ours (VP) W/O MKT&18.63& 11.53&8.12&16.42&12.02&6.31&4.57&12.66&\textbf{24.40}&\textbf{9.48}&2.62&18.53 \\ Ours (VP) &\textbf{28.24}&\textbf{14.68}&\textbf{9.51}&\textbf{21.45}&\textbf{12.36}&\textbf{6.47}&\textbf{4.92}&\textbf{13.48} &16.53&8.87&\textbf{4.53}&\textbf{20.19}\\ \hline \hline
				
				Ours (TP) W/O MKT &78.06&61.29&\textbf{43.87}&57.78&56.00&38.62&23.45&40.44&71.74&65.46&49.52&62.22 \\ 
				Ours (TP) &\textbf{80.65}&\textbf{61.94}&39.99&\textbf{58.32}&\textbf{60.14}&\textbf{43.59}&\textbf{28.28}&\textbf{43.48}&\textbf{78.74}&\textbf{69.81}&\textbf{53.14}&\textbf{65.74} \\
				\hline

		\end{tabular}}
		\vspace{-0.2cm}
		
		\label{t1}
					\vspace{-0.3cm}
	\end{table*}
					\vspace{-0.3cm}
	\subsection{Look-up Table}
	The outputs of different predictors are inconsistent shown in Fig.~\ref{sample2}. In order to solve the problem of semantic information deviation between cross-modalities, we design a Look-up Table $\mathbb{Q}$ to convert the output probabilities of one predictor as the target answer of another (such as converting the prediction subtitle timelines $T_{(s/e)}$ of the textual predictor to the corresponding frame span timepoints $V_{(s/e)}$, which realized information alignment of the cross-modal target answer.
	\begin{equation}
		\breve{T_s} = \mathrm{Argmin}\left( V_s-\mathbb{Q}(T_i)\right) ,\breve{T_e} = \mathrm{Argmin}\left( V_e-\mathbb{Q}(T_i)\right) 
	\end{equation}
	\begin{equation}
		\breve{V_s} = \mathrm{Argmin}\left( T_s-\mathbb{Q}(V_i)\right) ,\breve{V_e} = \mathrm{Argmin}\left( T_e-\mathbb{Q}(V_i)\right) 
	\end{equation}
	\subsection{Mutual Knowledge Transfer}
	In order to perform cross-modal mutual knowledge transfer, we introduce auxiliary objectives, and expect that the predictor can effectively learn the cross-modal information by predicting the output produced by the another-modal predictor. The whole procession is shown as follows, where the notation $\mathbb{E}$ means the Expectation. This represents that we want to transfer the knowledge as the pseudo label from one predictor to another.
	\begin{equation}
		\small
		\mathbb{E}([\mathbf{T_s^{Logits}},\mathbf{T_e^{Logits}}]) = [\breve{T_s},\breve{T_e}], ~\mathbb{E}([\mathbf{V_s^{Logits}},\mathbf{V_e^{Logits}}]) = [\breve{V_s},\breve{V_e}]
		\vspace{-0.2cm}
	\end{equation}
	\subsection{One-way Dynamic Loss Function}
	In the training stage, the cross-modal knowledge reserves for each predictor are different. We design a one-way dynamic loss function (ODL) that can adjust knowledge transferring. On the right of Fig.~\ref{framework}, ODL can dynamically adjust the proportion of knowledge transferring by comparing the matching between the prediction result and the target answer via the IoU function shown in equation (\ref{a1}).
	\begin{equation}
		\mathrm{IOU}(A,B) = \frac{A\cap B}{A \cup B}
		\label{a1}
	\end{equation}
\par
	Meanwhile, the knowledge difference between the textual predictor and the visual predictor will lead to inconsistent learning progress. Therefore, we use stop gradient (sg) for the two predictors to learn independently (this means one-way).
	\begin{equation}
		\small
		\mathrm{Loss_{Visual}^{Mutual}} = \alpha\times(\mathrm{CE}(\mathbf{V_s^{Logits}},sg(\breve{V_s}))+\mathrm{CE}(\mathbf{V_e^{Logits}},sg(\breve{V_e})))
	\end{equation}
	\begin{equation}
		\small
		\mathrm{Loss_{Textual}^{Mutual}} = \beta\times(\mathrm{CE}(\mathbf{T_s^{Logits}},sg(\breve{T_s}))+\mathrm{CE}(\mathbf{T_e^{Logits}},sg(\breve{T_e})))
	\end{equation}
where the $\alpha$ and $\beta$ can be calculated dynamically, which are presented as follows.
	\begin{equation}
		\small
		\alpha = IOU([\breve{V_s},\breve{V_e}],[V_s,V_e]),
		\beta = IOU([\breve{T_s},\breve{T_e}],[T_s,T_e])
	\end{equation}
	Finally, our loss function is:
	\begin{equation}
		\small
		\mathrm{Loss} = \mathrm{Loss_{Visual}}+\mathrm{Loss_{Textual}}+\mathrm{Loss_{Visual}^{Mutual}}+\mathrm{Loss_{Textual}^{Mutual}}
	\end{equation}
	\section{Experiment}
	\subsection{Experimental Setting}
	We evaluate MutualSL in three different public VAL datasets, where these datasets are formed with the text questions and corresponding visual answer clips as the target answers. The MedVidQA \cite{gupta2022dataset} is a medical instructional dataset that contains 3,010 question-and-answer (QA) pairs and 899 videos; TutorialVQA \cite{AnthonyColas2019TutorialVQAQA} contains 76 tutorial videos about software editing tutorials with 6,195 QA pairs; The VehicleVQA \cite{HongyinLuo2019IntegratingVR} dataset has a series of \textit{How-To} videos that introduce practical instructions on vehicles, including 9,482 QA pairs within 107 videos. Following previous works \cite{gupta2022dataset, MengLiu2018AttentiveMR,YitianYuan2019ToFW}, we use IoU-0.3/0.5/0.7 and mIoU as the evaluation metrics to compare several state-of-the-art (SOTA) methods on VAL tasks. We use the same visual extractor and text extractor in each baseline to ensure fairness, and follow the original author's parameter settings. We compare with several SOTA methods, VSLBase/VSLNet \cite{zhang2020span}, TMLGA \cite{CristianRodriguezOpazo2020ProposalfreeTM}, ACRM \cite{tang2021frame} and RaNet \cite{JialinGao2021RelationawareVR} use visual predictor to predict frame span timepoint. MoR \cite{kusa-etal-2022-dossier} and VPTSL \cite{li2022towards} use a textual predictor to predict textual subtitle span, which are the competitive SOTA methods.
	In the parameter settings of MutualSL, we set $d$ = 1024 and use the AdamW optimizer \cite{IlyaLoshchilov2017DecoupledWD}, where lr = 1e-5. We use Pytorch in three A100 GPUs for experiments, where the batch size = 4 and training epoch = 15. For all experiments, we repeat three-time experiments to reduce the random errors.
	\begin{figure}[t]
		\centering
		\includegraphics[width=8cm]{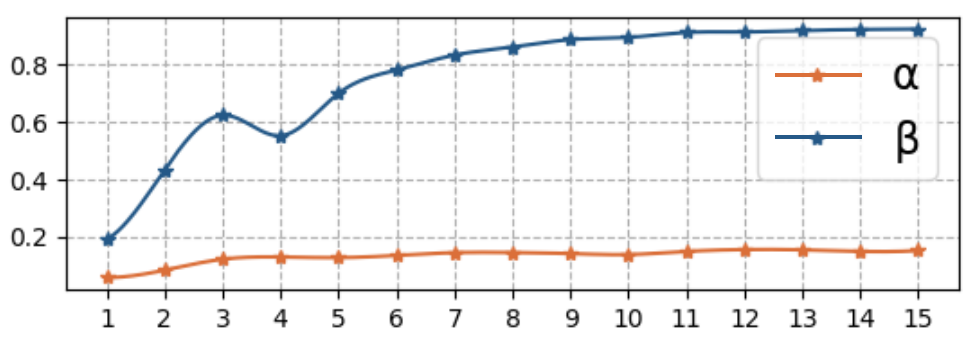}
		\vspace{-0.2cm}
		\caption{The changing trend of $\alpha$ and $\beta$ during training.}
		\label{framework1}
		\vspace{-0.5cm}
	\end{figure}
			\vspace{-0.6cm}
	\subsection{Results}

	As shown in Table \ref{t11}, we compare the performance of different methods in three datasets of the VAL task. The MutualSL achieves SOTA performance in most metrics, which shows the effectiveness of our method, especially the mIoU increases by 0.51, 3.28, and 1.23 respectively. The reason may be that mutual knowledge transfer (MKT) can guide the predictor to understand different information, thus alleviating the deviations of knowledge in different modalities. 
	\par
	To further analyze the impact of MKT on the ODL under different predictors, we perform the ablation study of hyper-parameters $\alpha$ and $\beta$ shown in Fig.~\ref{framework1}. We can clearly see that both $\alpha$ and $\beta$ increase with the increase of the epoch. The $\beta$ is more than $\alpha$, because the textual predictor has better answering localization ability, which is also in line with the experimental results shown in Table \ref{t11}. The Fig.~\ref{framework1} also shows that the ODL can dynamically adapt the ability of knowledge transfer from different predictors. 
	\par We’ve conducted extensive ablation experiments to analyze the MKT in Table \ref{t1}. In the visual predictor, using MKT can improve the mIoU indicators of the three datasets by 0.69 on average; The average increase in the textual predictor is 2.33 mIoU. These prove that the use of MKT can enhance the model's perception of different modal information, thus improving the performance of VAL. We find that although the improvement of the visual predictor is low, the MKT can greatly improve the performance of the textual predictor. Meanwhile, the performance of the textual predictor outperforms the visual predictor. Therefore, we use the result of the textual predictor as the output of MusicalSL in the prediction phase.
			\vspace{-0.35cm}
	\section{Conclusion}
		\vspace{-0.1cm}
	In this paper, we proposed a cross-modal mutual knowledge transfer method (MutualSL) for VAL tasks. This method alleviates the problem of knowledge deviation, which uses visual predictor and textual predictor for cross-modal mutual knowledge transfer. We compare and ablate the proposed methods in three public datasets of the VAL task, where the proposed method outperforms all competitive SOTA methods. This proves the effectiveness of the MutualSL. In the future, we hope to explore more methods such as knowledge distillation for understanding cross-modal knowledge to promote the development of related fields.

	\clearpage
	\bibliographystyle{IEEEbib}
	\bibliography{strings,refs}
	
\end{document}